\documentclass[letterpaper,10pt, conference]{ieeeconf} 

\IEEEoverridecommandlockouts  
\usepackage[pdftex]{graphicx}
\usepackage{epsfig} 
\usepackage{times} 
\usepackage{amsmath} 
\usepackage{amssymb}  
\usepackage{url}
\usepackage{hyperref}

\usepackage{comment}
\usepackage{mathtools}
\usepackage[font=footnotesize]{caption}
\usepackage[subrefformat=parens,font=footnotesize]{subcaption}
\usepackage[utf8]{inputenc}

\usepackage[ruled,linesnumbered,noend]{algorithm2e}
\SetAlCapFnt{\footnotesize}
\usepackage{footmisc}

\makeatletter
\newcommand{\figcaption}[1]{\def\@captype{figure}\caption{#1}}
\newcommand{\tblcaption}[1]{\def\@captype{table}\caption{#1}}
\makeatother

\newcommand{\argmin}{\mathop{\rm arg~min}\limits}
\newcommand{\figurename}{Fig.~} 
\newcommand{\equationname}{Eq.~}
\newcommand{\tablename}{TABLE~}

\usepackage{booktabs}
\usepackage{longtable}
\usepackage{color}

\title{\LARGE \bf Learning to Shape by Grinding:\\Cutting-surface-aware Model-based Reinforcement Learning}

\author{Takumi Hachimine$^{1}$, Jun Morimoto$^{2,3}$, and Takamitsu Matsubara$^{1}$
\thanks{This work was supported by JST-Mirai Program Grant Number~JPMJMI21B1, Japan.}
\thanks{$^{1}$T. Hachimine~and~T. Matsubara are with
the Graduate School of Information Science, Nara Institute of Science and
Technology (NAIST), Nara, Japan.}%
\thanks{$^{2}$J. Morimoto is with the Department of Systems Science, Graduate
School of Informatics, Kyoto University, Kyoto, Japan.}%
\thanks{$^{3}$J. Morimoto is also with the Brain Information Communication Research Laboratory Group (BICR), Advanced Telecommunications Research
Institute International (ATR), Kyoto, Japan.}%
}

\begin{document}
\maketitle
\thispagestyle{empty}
\pagestyle{empty}

\begin{abstract}
Object shaping by grinding is a crucial industrial process in which a rotating grinding belt removes material.
Object-shape transition models are essential to achieving automation by robots; however, learning such a complex model that depends on process conditions is challenging 
because it requires a significant amount of data, and the irreversible nature of the removal process makes data collection expensive. 
This paper proposes a \textit{cutting-surface-aware} Model-Based Reinforcement Learning (MBRL) method for robotic grinding. 
Our method employs a cutting-surface-aware model as the object's shape transition model, which in turn is composed of a geometric cutting model and a cutting-surface-deviation model, based on the assumption that the robot action can specify the cutting surface made by the tool.
Furthermore, according to the grinding resistance theory, the cutting-surface-deviation model does not require raw shape information, making the model's dimensions smaller and easier to learn than a naive shape transition model directly mapping the shapes. 
Through evaluation and comparison by simulation and real robot experiments, we confirm that our MBRL method can achieve high data efficiency for learning object shaping by grinding and also provide generalization capability for initial and target shapes that differ from the training data.
\end{abstract}

\section{INTRODUCTION}
\begin{figure}[t]
    \centering
    \includegraphics[clip,width=1.0\columnwidth]{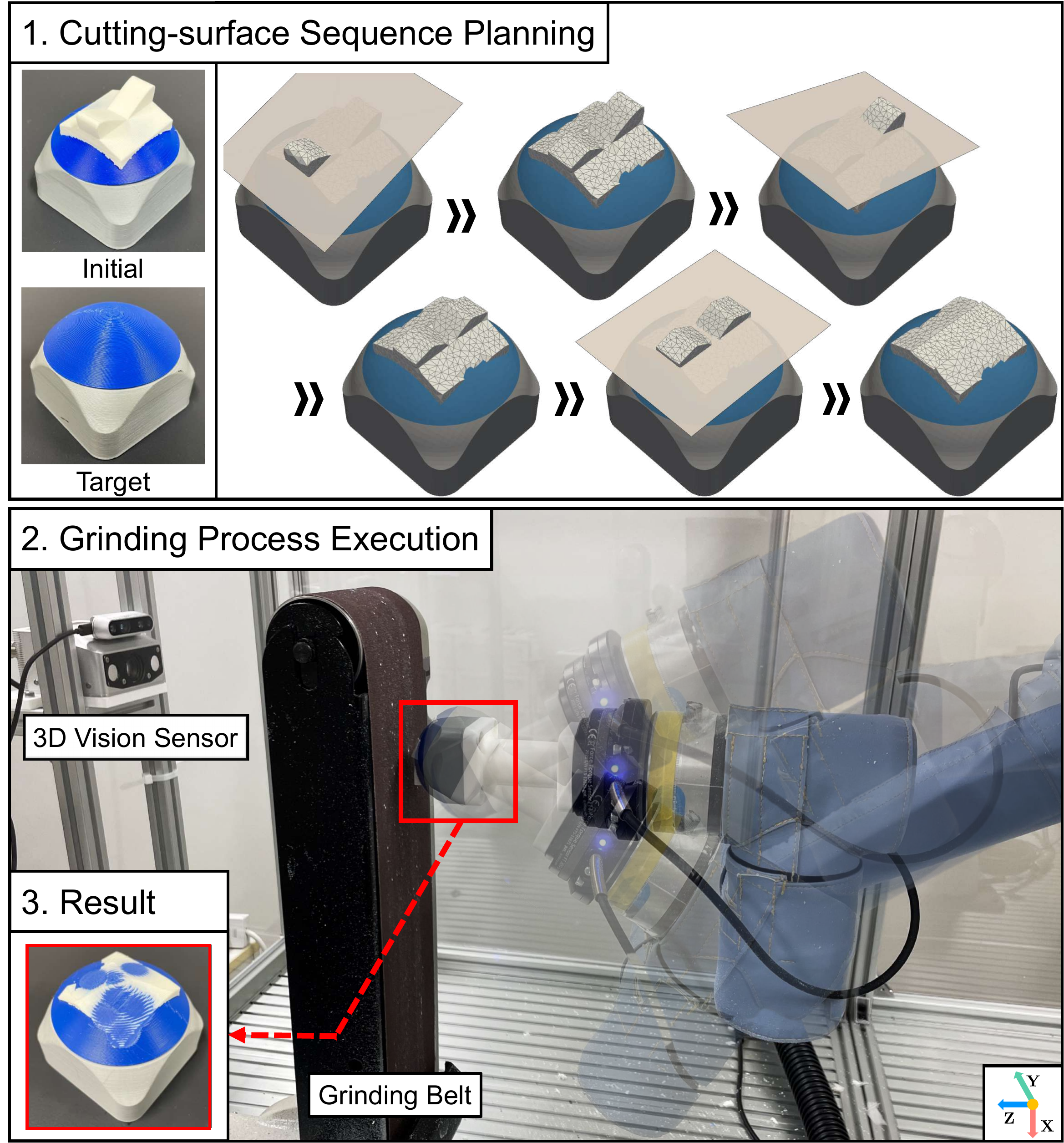}
    \caption{{Robotic object shaping by grinding}.
    The robot shapes the object from the initial to the target shape, considering shape deviation by removal resistance.
    Robot actions are planned as a sequence of the cutting surface.}
    \label{fig:eye_cath}
    \vspace{-4.0truemm}
\end{figure}

Object-shape manipulation is widely applied in industry and our daily life. 
Shaping methods include two major types: Deformation processing \cite{sanchez2018robotic} shapes objects by bending or pressing, in the manner of forming a dough, while removal processing shapes objects by gradually removing unnecessary material from the base stock, such as cutting metal. 
As a common industrial removal process, this paper focuses on the grinding process shown in \figurename\ref{fig:eye_cath}, in which material is removed by a rotating grinding belt, and explores the possibility of automating this process with a robot.

To enable the grinding process by a robot, modeling the object-shape transition is crucial; however, 
this is difficult because the removal process generates removal resistance depending on material hardness, abrasive grain size, and belt rotation speed \cite{9341102}, which drags on the robot's end-effector and complicates the shape transition. 
Even if such environmental conditions are fixed, different processing positions and postures of the robot generate different removal resistance, which affects removal shape and volume and thus still complicates the shape transition. Therefore, in this paper, as a first step toward automating the grinding process, we focus on the grinding process with various processing positions and postures of the robot as the process condition, while the environmental conditions are fixed. 

Due to the complexity of the shape transition model in removal processes, it is reasonable to consider a data-driven model learning approach. Model-Based Reinforcement Learning (MBRL) has been successfully applied to various tasks as a framework that efficiently performs model learning and control planning \cite{matl2021deformable, nagabandi2020deep,deisenroth2011pilco}.
In particular, we focus on the PILCO framework \cite{deisenroth2011pilco}, which uses Gaussian process regression for model training. This is because it allows for the evaluation of the predictive uncertainty of the model, which is useful for grinding tasks that require avoiding robot motions with a high risk of damage due to model uncertainty.
A naive model that takes the current shape and process conditions as inputs and learns a mapping that outputs the shape at the next step is, however, problematic because the shape representation (point cloud, depth image, etc.) is very high-dimensional and the shape transition depends on the process conditions. 
Therefore, a model learned with one particular initial and target shape cannot apply to different initial and target shapes, even for the same material is prone to generalization error.
Therefore, an MBRL framework that satisfies 1) high data efficiency and 2) applicability of the learned models to different initial and target shapes is needed for autonomous object shaping by grinding.

To satisfy the above requirements, the removal process is performed by local surface contact between the tool surface (grinding belt) and the object. In other words, by interpreting that the shape transition is accomplished by the {\it cutting surface}, the shape transition can be represented by a) a geometric shape transition model with a cutting surface that ignores the effects of process conditions and b) a model that predicts such cutting-surface deviation that is dependent on process conditions.
The former model can be commonly applied to a wide variety of shapes because the transitions are geometrically computable and require no learning.
The latter model needs learning, but according to the grinding resistance theory \cite{TANG20092847}, it does not require raw shape information, which reduces model dimensions and generalization capability. 

With the above in mind, we propose a cutting-surface-aware MBRL method for object shaping by grinding. 
Our shape transition model, {Cutting-Surface-Aware Model (CSAM)}, is composed of a) {a Geometric Cutting Model} (GCM) as a geometric shape transition model and b) {a Cutting-Surface-Deviation Model (CSDM)} that predicts the deviation of the cutting surface depending on the process conditions, based on the assumption that the robot action can specify the cutting surface made by the tool.
Our MBRL iteratively optimizes both CSAM through learning CSDM and a cutting-surface sequence as robot actions through {Model Predictive Control} (MPC) for object shaping by grinding.
We verified the performance of the proposed method in simulation and with real robot experiments.
The results indicate that our method can achieve high data efficiency for learning object shaping by grinding and also provide {generalization capability for initial and target shapes that differ from the training data.}
The following are this paper's main contributions:
\begin{itemize}
    \item We proposed the first shape transition modeling method based on the cutting surface for grinding processes by a robot, taking into account the high dimensionality of the shape representation and dependence on process conditions. 
    \item We proposed an MBRL method for learning a shape transition model and planning the optimal cutting-surface sequence with MPC.
    \item We verified the performance of the proposed method in simulation and real robot experiments. 
\end{itemize}

\section{RELATED WORK}
\subsection{Robotic Object-shape Manipulation}
There have been many studies on shape manipulation tasks using robots. 
However, conventional methods focus on the deformation process, such as ropes \cite{8403315,9197121}, cables \cite{9001169,9561391}, and fabrics \cite{corl2020softgym,TSURUMINE201972}.
In contrast, studies on the removal process have been limited, perhaps due to the difficulty of modeling the shape transition depending on the process conditions.
Methods for shape splitting have been studied in the Computer Vision field \cite{inproceedings_meshcut}, but limited to geometrical computation algorithms.
A related study of removal processing is cutting food items with a knife.
Heiden \textit{et al}. developed a differentiable cutting simulator to reproduce the force applied to a knife when cutting food \cite{heiden2021disect}.
By optimizing simulation parameters based on actual measurements, they constructed a simulator for various ingredients.
Another related study involved polishing tasks.
In order to improve surface properties by polishing, Yang \textit{et al}. proposed a framework that collects process conditions to obtain the optimal parameters that satisfy the desired surface property using a genetic algorithm \cite{9341102}.

In any case, these studies did not construct a shape transition model due to the complexity of the shape transition by the removal process.
Therefore, the conventional method is an adaptive method that optimizes parameters ({e.g.,} robot speed and material properties) within an arbitrarily defined trajectory.
In contrast, our proposed method can model the shape transition based on the cutting surface and plan robot action using MPC.

\subsection{Object-shape Manipulation with Reinforcement Learning}
Several studies have been conducted using both model-free RL and model-based RL. These methods may be preferable for grinding processes because there is no risk of injury or physical burden on the human demonstrator compared to such imitation learning approaches as Behavior Cloning \cite{pomerleau1991efficient} that use demonstration data. Model-free RL is effective for learning policy directly from high-dimensional inputs; however, it requires a large amount of data collection to obtain optimal control policies.
Physics simulators for deformation processing \cite{corl2020softgym,hu2019difftaichi} are often used to collect training data because real-world experiments are costly. 
A typical physics simulator for removal processes is {Computer-Aided Manufacturing} (CAM).
CAM can calculate the removal resistance applied to the tool and the processing time by setting variables such as the tool path and processing parameters. 
Therefore, a very high-power and high-rigidity mechanism is required, which is not suitable for shape manipulation with a multi Degree of Freedom (DoF) robot.
Furthermore, the finite element and particle methods are computationally expensive.

Consequently, we employ the MBRL framework for high data efficiency.
In addition, by introducing CSAM, we only need to learn CSDM, which depends on process conditions, instead of the direct learning input-output shape-transition model.
This approach allows certain generalization capabilities for different initial and target shapes with a reasonable amount of data collection.

\section{PROPOSED METHOD}
This section explains our proposed method. 
Section \ref{subsection:cutting_surface_aware_transition_model} describes the shape transition model called CSAM,
Section \ref{subsection:cutting_surface_aware_MPC} discusses the planning of cutting-surface sequences by MPC, and Section  \ref{subsection:cutting_surface_aware_MBRL} introduces our MBRL method that iteratively optimizes the CSDM and action plans.

\begin{figure}[t]
    \centering
    \includegraphics[width=1.0\columnwidth]{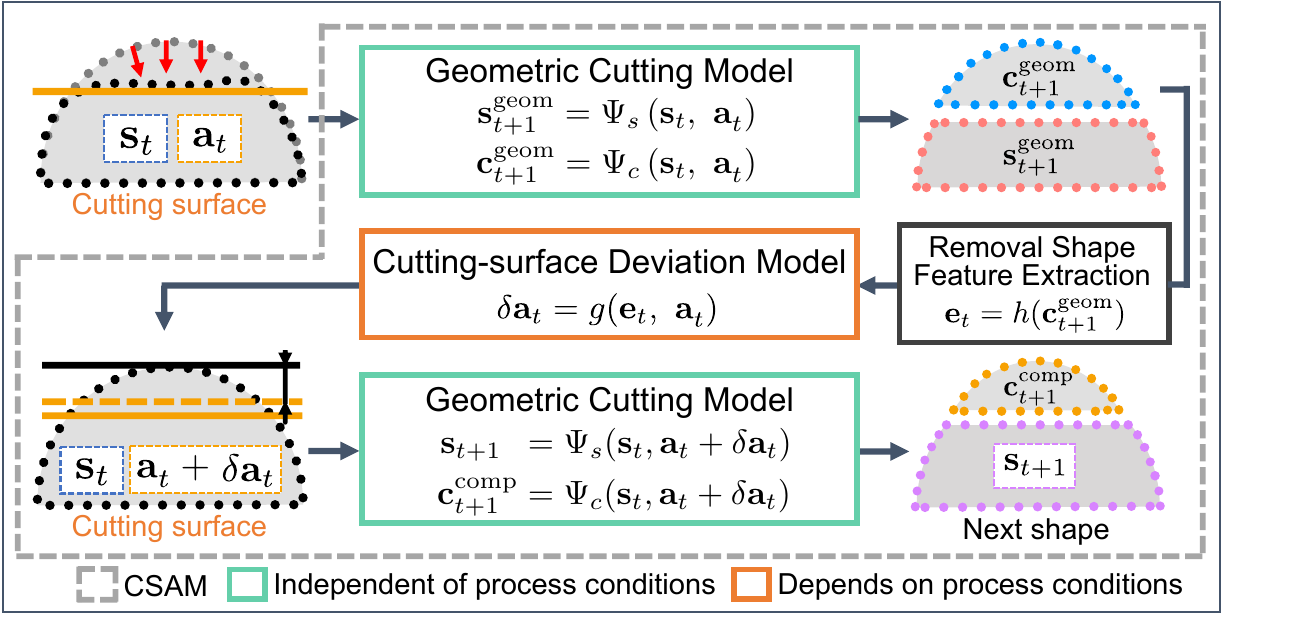}
    \caption{Representation of removal process shape-transition model by {Cutting-Surface-Aware Model (CSAM)}, which is composed of a) a {Geometric Cutting Model (GCM)} for geometrically splitting the shape by the cutting surface independent of the process conditions and b) a {Cutting-Surface-Deviation Model (CSDM)} for predicting the deviation of the cutting surface due to removal resistance.}
    \label{fig:propose_method_large:b}
    \vspace{-2.8truemm}
\end{figure}

\subsection{Cutting-Surface-Aware Model (CSAM)}\label{subsection:cutting_surface_aware_transition_model}
The proposed CSAM shown in \figurename\ref{fig:propose_method_large:b} is composed of a) GCM, which geometrically splits the shape by the cutting surface independent of the process conditions, and b) CSDM, which predicts the deviation of the cutting surface due to removal resistance.
We denote \(\mathbf{s}_t \in \mathcal{S}\) as the state representing the object shape, \(\mathbf{a}_t \in \mathcal{A}\) as the robot action, and \(t\) as the time step. 
In our experiments as shown later, we employed the point cloud \(\mathcal{O}_t:=[{\mathbf q}_{i,t}]_{i=1}^{D}\subseteq \mathbb{R}^{3}\) as the state, where \(D\) is the number of points and \(\mathbf{q}_{i,t}\) is the \(i\)-th particle position. 
Here, we interpret the shape transition as caused by the cutting surface; therefore, the robot's action is defined as a specification of the cutting surface. 

\begin{figure}[t]
    \centering
    \includegraphics[clip,width=0.85\columnwidth]{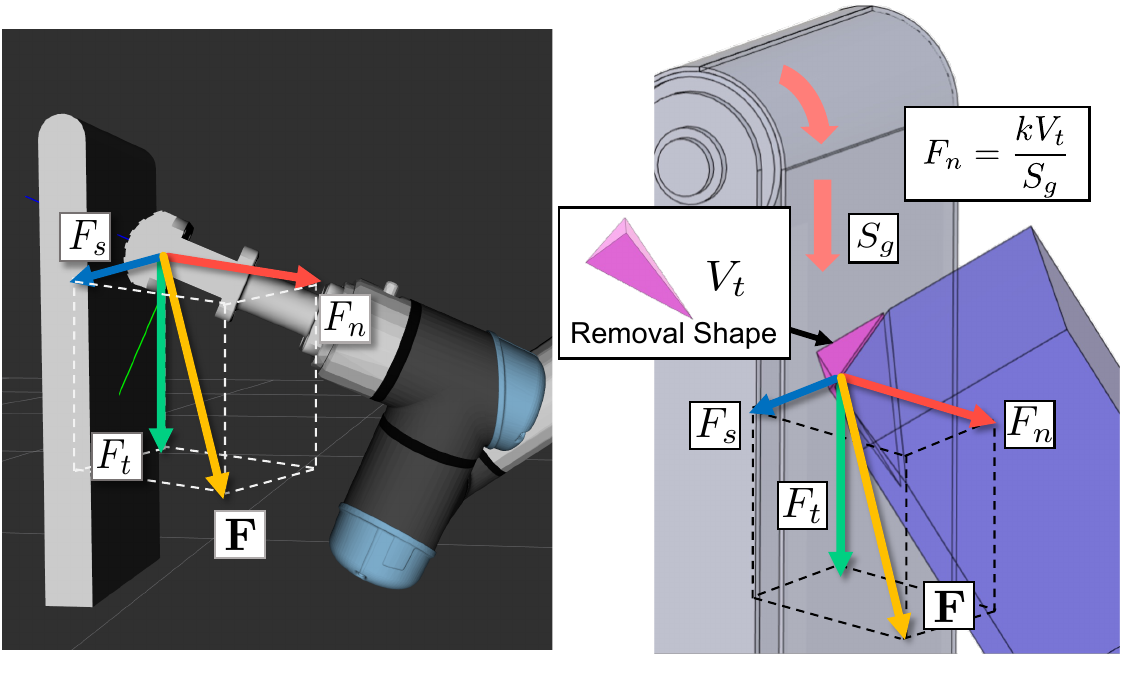}
    \caption{Three component forces of grinding resistance on grinding belt}
    \label{fig:grinding_resistance}
    \vspace{-2.8truemm}
\end{figure}

\subsubsection{Geometric Cutting Model (GCM)}
We assume that a cutting surface (robot action) \(\mathbf{a}_t\) geometrically splits the current shape \({\mathbf{s}}_{t}\) into the next time-step shape \({\mathbf{s}}_{t+1}^{\rm geom}\) and removal shape \({\mathbf{c}}_{t+1}^{\rm geom}\).
Shape deformation by the cutting surface is defined as the following geometric
cutting functions \(\Psi(\cdot)\):
\begin{align}
    {\mathbf s}_{t+1}^{\rm geom}= {\Psi_s}\left(\mathbf{s}_t,~{\mathbf{a}}_t\right), 
    {\mathbf c}_{t+1}^{\rm geom}= {\Psi_c}\left(\mathbf{s}_t,~{\mathbf{a}}_t\right).
    \label{eq:cutting_function}
\end{align}
These functions only geometrically split the shape by the cutting surface. 
Shape splitting is tractable because it is based on collision detection between the cutting surface and the shape \cite{kumaresan2000linear,dunn20113d}, in which computational cost is linear to the number of point clouds \(D\).

\subsubsection{Cutting-Surface Deviation Model (CSDM)}
In actual removal processing, removal resistance occurs between the robot and the object, depending on the process conditions.
In particular, the removal resistance generated by the grinding is called grinding resistance \(\mathbf{F}=[F_t, F_n, F_s]^\mathsf{T}\) as illustrated in \figurename\ref{fig:grinding_resistance}. 
This resistance operates in 1) a tangential direction to the grinding belt \(F_t\), 2) a vertical direction to the grinding belt surface \(F_n\), and 3) parallel directions to the circumference of the grinding belt \(F_s\). 

Although the analytical formulation of grinding resistance is difficult due to the complexity of the physical grinding process, experimental results show that \(F_n = k {{V}_{t}}/{S_g}\) and \(F_t = \lambda k {{V}_{t}}/{S_g}\) \cite{TANG20092847}, \cite{rowe2013principles}.
Both \(F_n\) and \(F_t\) are proportional to the removal volume \({V}_{t}\) and inversely proportional to the rotational speed of the grinding belt \(S_g\), where \(k\) is the specific grinding energy and \(\lambda\) is the ratio of \(F_n/F_t\). 
In practice, the grinding resistance does not match the experimental formula owing to various other effects (e.g., friction heat, clogging of the grinding belt, the other force \(F_s\)).

With this grinding resistance theory \(F_n = k {{V}_{t}}/{S_g}\), \(F_t = \lambda k {{V}_{t}}/{S_g}\), we construct CSDM \(g(\cdot)\), using a data-driven approach, to predict the deviation of the cutting surface due to grinding resistance, which depends on process conditions. 
The model takes the removal shape features \(\mathbf{e}_t=h({{\mathbf c}_{t+1}^{\rm geom}})\) and the robot action \(\mathbf{a}_t\) (cutting surface) as input. \(h(\cdot)\) is the function that extracts features (e.g., volume, size) from the removal shape \({\mathbf c}_{t+1}^{\rm geom}\) based on grinding resistance theory.
By using the CSDM \(g(\cdot)\), the deviation of the cutting surface \(\delta \mathbf{a}_t\) can be expressed as follows:
\begin{align}
    {\delta {\mathbf a}_t}= g({\mathbf e}_t,\mathbf{a}_t), \mathbf{e}_t=h\left({{\mathbf c}_{t+1}^{\rm geom}}\right).
    \label{eq:residual_error_function}
\end{align}
The deformed shape \(\mathbf{s}_{t+1}\) can be expressed using the geometric cutting function \(\Psi_s(\cdot)\) and the cutting-surface deviation function \(g(\cdot)\):
\begin{equation}
\begin{split}
\mathbf{s}_{t+1}&={\Psi_s}\left(\mathbf{s}_t,\mathbf{a}_t+{\delta {\mathbf a}_t}\right)={\Psi_s}\left(\mathbf{s}_t,\mathbf{a}_t+g\left(\mathbf{e}_t,\mathbf{a}_t\right)\right)\\
&={\Psi_s}\left(\mathbf{s}_t,\mathbf{a}_t+g\left(h\left(\Psi_c\left(\mathbf{s}_t, \mathbf{a}_t\right)\right),\mathbf{a}_t\right)\right)\\
&\triangleq f\left(\mathbf{s}_{t},\mathbf{a}_t\right).
\end{split}
\end{equation}
In summary, by representing the shape-transition model as GCM and CSDM, it can be expressed by the definition of the cutting surface at each time step. 
Note that CSDM does not use the removal shape directly but only the extracted features. 
Therefore, the learned CSDM is expected to have the ability to generalize across different initial and target shapes.

\subsubsection{CSDM training with Gaussian Process}
In this study, Gaussian process (GP) regression \cite{3569} is used to learn the CSDM. 
The deviation of cutting surface \(\delta{\mathbf {a}_t}\) depends on the extracted features of the removal shape \(\mathbf{e}_t\) and robot action \(\mathbf{a}_t^{}\). 
Thus, We define \(\mathbf{{X}}=\left[\mathbf{x}_{1},\dots,\mathbf{x}_{N}\right]^\mathsf{T}\), \({\mathbf x}_n:=\left[{\mathbf e}_t,{\mathbf a}_t^{}\right]\) as training inputs, and \(\mathbf{{Y}}=\left[{\delta a}_{1},\dots,{\delta a}_{N}\right]^\mathsf{T}\) as training outputs, where \(N\) is data size.
 
Assuming the GP as prior distribution for the function \(g(\cdot)\), the predicted mean \(m({\mathbf{x}}_{*})\) and predicted variance \(\sigma^{2}({\mathbf{x}}_{*})\) for the new input data \({\mathbf{x}}_{*}\) are computed as follows:
\(p\left(g({\mathbf{x}}_{*})\mid{\mathbf{X}},{\mathbf{Y}}\right) = \mathcal{N}\left(g_{}({\mathbf{x}}_{*})\mid{m}({\mathbf{x}}_{*}), \sigma^{2}_{_{}}({\mathbf{x}}_{*})\right), m_{_{}}({\mathbf{x}}_{*}) = \mathbf{k}_{*}^{\mathsf{T}}(\mathbf{K}^{}+\sigma^{2}_{_{}}\mathbf{I})^{-1}\mathbf{Y}^{}, \sigma^{2}_{_{}}({\mathbf{x}}_{*}) = k_{**} - \mathbf{k}_{*}^{\mathsf{T}}(\mathbf{K}^{}+\sigma^{2}_{_{}}\mathbf{I})^{-1}\mathbf{k}_{*},\)
where \({\mathbf{k}}_{*}=k({\mathbf{X}},{\mathbf{x}}_{*})\),  \(k_{**}=k({\mathbf{x}}_{*},{\mathbf{x}}_{*})\), \(k(\cdot,\cdot)\) is kernel function, and \(\mathbf{K}\) is a kernel gram matrix that has \(K_{i,j}^{}=k_{}(\tilde{\mathbf{x}}_{i}, \tilde{\mathbf{x}}_{j})\).
Therefore, the deviation of cutting surface \(\delta {a}_{*}\) is predicted as GP mean \(m_{_{}}({\mathbf{x}}_{*})\), and \(\sigma^{2}_{_{}}({\mathbf{x}}_{*})\) is the variance of \(\delta {a}_{*}\), which is used as the cost function. 
Grinding resistance theory suggests removal resistance is proportional to removal volume at one time; hence, such a smooth kernel as Squared Exponential (SE) kernel can be used.
If the output values are multidimensional, the prediction is executed by preparing multiple GPs for the number of dimensions.

\subsection{Cutting-surface Sequence Planning with MPC}\label{subsection:cutting_surface_aware_MPC}
With the learned shape-transition model, the optimal robot-action sequence from the current shape to the target shape can be obtained using an MPC framework, which can be formulated as follows:
\begin{equation}
 \left.
 \begin{aligned}
\left[\mathbf{a}^{*}_{t+1},\dots,\mathbf{a}^{*}_{t+H}\right] &= \argmin_{\mathbf{a}_{t+1},\ldots,\mathbf{a}_{t+H}}\frac{1}{H}{\sum^{t+H}_{\tau=t+1}}C(\hat{\mathbf{s}}_{\tau}, \mathbf{a}_{\tau}),\\
  \text{subject to}~\mathbf{s}_{t+1}&=f(\mathbf{s}_{t},\mathbf{a}_t),
 \end{aligned}
 \right\}
\label{eq:mpc_objective_function}
\end{equation}
where \(H\) is the planning horizon, \(\tau\) is the planning index, \(t=1,2,\ldots,T\) is the task horizon, and \(\hat{\mathbf{s}}_{\tau}\) is the shape predicted by CSAM.
The cost function \(C(\hat{\mathbf{s}}_\tau,\mathbf{a}_\tau)\) for object shaping can be defined as:
\begin{equation}
C(\hat{\mathbf{s}}_{\tau},{\mathbf a}_{\tau})=
\begin{aligned}[t]
&\underbrace{d_{\rm CD}(\hat{\mathbf{s}}_{\tau},\mathbf{s}_{\rm{target}})}_{\rm {Shape~error~cost}}+\underbrace{\phi(\hat{\mathbf{s}}_{\tau},\mathbf{s}_{\rm{target}})}_{\rm Over~cutting ~cost}\\
&+\eta \underbrace{{\frac{1}{M}\sum_{j=1}^{M}}\sigma^{2}_{j}\left({\left[h\left(\Psi_c\left(\hat{\mathbf{s}}_{\tau}, \mathbf{a}_{\tau}\right)\right),{\mathbf a}_{\tau}\right]}_{}\right)}_{{\rm Variance~cost}},
\label{eq:mpc_cost_func}
\end{aligned}
\end{equation}
where \(\left[h\left(\Psi_c\left(\hat{\mathbf{s}}_{\tau}, \mathbf{a}_{\tau}\right)\right),{\mathbf a}_{\tau}\right]=[{\mathbf e}_{\tau},{\mathbf a}_{\tau}^{}]={\mathbf{x}}_{*}\), \(d_{\rm CD}\left({\mathbf{s}_{1}, \mathbf{s}_{2}}\right)\) is Chamfer discrepancy \cite{Nguyen2021PointsetDF}, which evaluates the error between two point-cloud data sets:
\begin{equation}
d_{\rm CD}(\mathbf{s}_1,\mathbf{s}_2)=\frac{1}{|\mathbf{s}_1|}
\sum_{\mathbf{p} \in \mathbf{s}_1}\underset {\mathbf{q} \in \mathbf{s}_2}{\rm min} \|\mathbf{p}-\mathbf{q}\|_{2}^{2}+\frac{1}{|\mathbf{s}_2|}
\sum_{\mathbf{q} \in \mathbf{s}_2}\underset {\mathbf{p} \in \mathbf{s}_1}{\rm min} \|\mathbf{p}-\mathbf{q}\|_{2}^{2}.
\label{eq:champer_distaces}
\end{equation}

In \equationname\ref{eq:mpc_cost_func}, the shape error cost refers to error between the target shape \(\mathbf{s}_{\rm{target}}\) and the predicted shape \(\hat{\mathbf{s}}_{{\tau}}\).
The over-cutting cost is a penalty for the state of over-cutting the target shape.
The variance cost refers to avoiding uncertain actions, \(\sigma^{2}_{j}(\cdot)\) is the predicted variance of the GP, \(M\) is the number of predicted dimensions, and \(\eta\) is the coefficient.

\begin{figure}[t]
    \centering
    \includegraphics[clip,width=1.0\columnwidth]{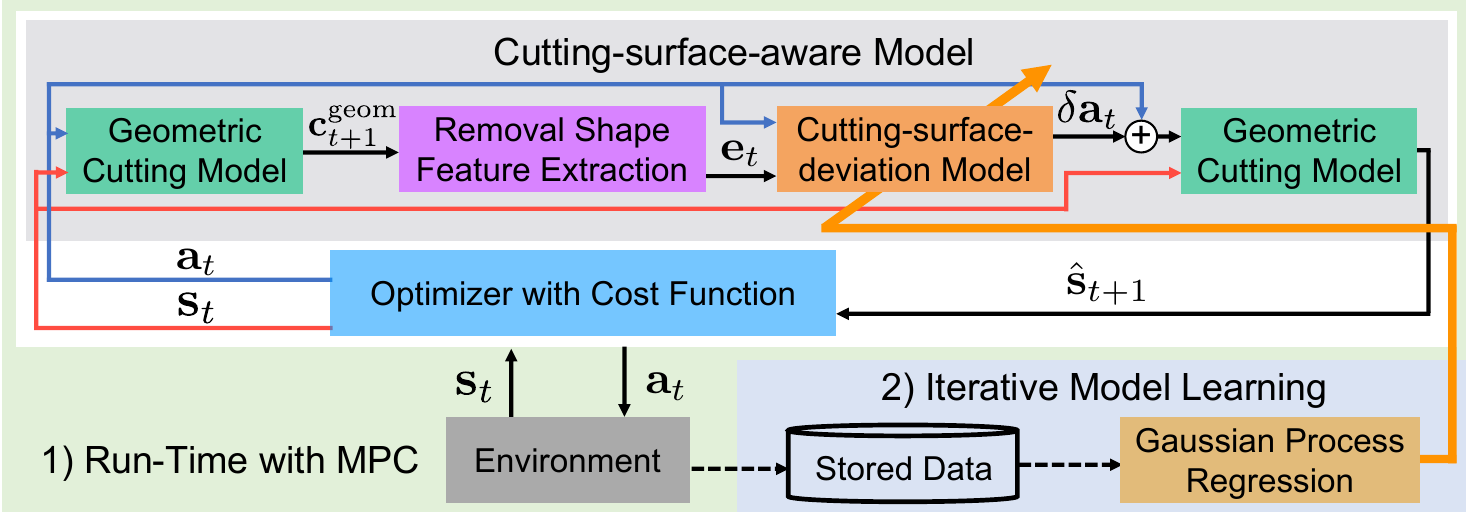}
    \caption{Our MBRL framework of the object-shaping planner for the removal process.
        The proposed method iteratively optimizes 1) the planning (and execution) of the cutting-surface sequences by MPC using the cutting-surface-aware model and 2) the training and updating of the cutting-surface-deviation model with collected data from MPC run-time. }
    \label{fig:propose_method_large:c}
    \vspace{-2.8truemm}
\end{figure}

\begin{algorithm}[!t]
\footnotesize
    \caption{\footnotesize Cutting-surface Aware MBRL}\label{algo:mbrl_alg}
        \SetKwFunction{Reset}{Set New Object}
        \SetKwFunction{RS}{Get state}
        \SetKwFunction{RA}{RandomActions}
        \SetKwFunction{AS}{ActionSearch}
        \SetKwFunction{CMD}{SetCommand}
        \SetKwFunction{EX}{Execute}
        \SetKwFunction{CF}{costFunction}
        \SetKwFunction{train}{train CSDM}
        \SetKwFunction{CSAM}{CSAM}
        \SetKwFunction{dyModel}{model}
        \SetKwComment{Comment}{$\triangleright$}{}
  \ \textbf{{Input}}: RL episode $N_{episode}$, Task horizon $T$, Predict Horizon $H$, Cost function ${C}(\cdot)$\\
  \ \textbf{{Initialization}}: CSDM dataset $\mathcal{D}\leftarrow\{\mathbf{X} = \left\{ \right\}$,~$\mathbf{Y} = \left\{ \right\}\}$ \\
  \# Generate random samples with random actions\\
  \Reset{}\\
  \For{$ i = 1,...,N_{init}$}
  {
    $\mathbf{s}_i$ = \RS{}\\
    $\mathbf{a}_i$ = \RA{}\\
    \EX{$\mathbf{a}_i$}\\
    $\mathbf{s}_{i+1}$ = \RS{}\\
    $\mathbf{x}_{*}\leftarrow [{\mathbf e}_{i},{\mathbf a}_{i}^{}]$, $\mathbf{y}_{*}\leftarrow [{\delta {\mathbf a}}_{i}]$\\
    $\mathbf{X} = \{\mathbf{x},{\mathbf x}_{*}\}, \mathbf{Y} = \{\mathbf{y}, \mathbf{y}_{*}\}$
  }
  \dyModel = \train{$\mathcal{D}$}\\
  \# Model-based RL with MPC\\
  \For(\Comment*[f]{{\footnotesize Iterative Model Learning}}){$i = 1,...,N_{episode}$}{
    \Reset{}\\
    \For(\Comment*[f]{{\footnotesize Run-Time with MPC}}){$t = 1,...,T$}{
      $\mathbf{s}_{t}$ = \RS{}\\
      $\mathbf{a}^{}_{t}$ = \AS{${\mathbf{s}}_t, H, \CSAM,~{C}(\cdot)$}\\
      \EX{$\mathbf{a}^{}_{t}$}\\
      $\mathbf{s}_{t+1}$ = \RS{}\\
      $\mathbf{x}_{*}\leftarrow [{\mathbf e}_{t},{\mathbf a}_{t}^{}]$, $\mathbf{y}_{*}\leftarrow [{\delta {\mathbf a}}_{t}]$\\
      $\mathbf{X} = \{\mathbf{x},\mathbf{x}_{*}\}, \mathbf{Y} = \{\mathbf{y},\mathbf{y}_{*}\}$
    }
    \dyModel = \train{$\mathcal{D}$}\\
  }
\end{algorithm}

\begin{figure}[t!]
    \centering
    \vspace{-1truemm}
    \includegraphics[clip,width=0.75\columnwidth]{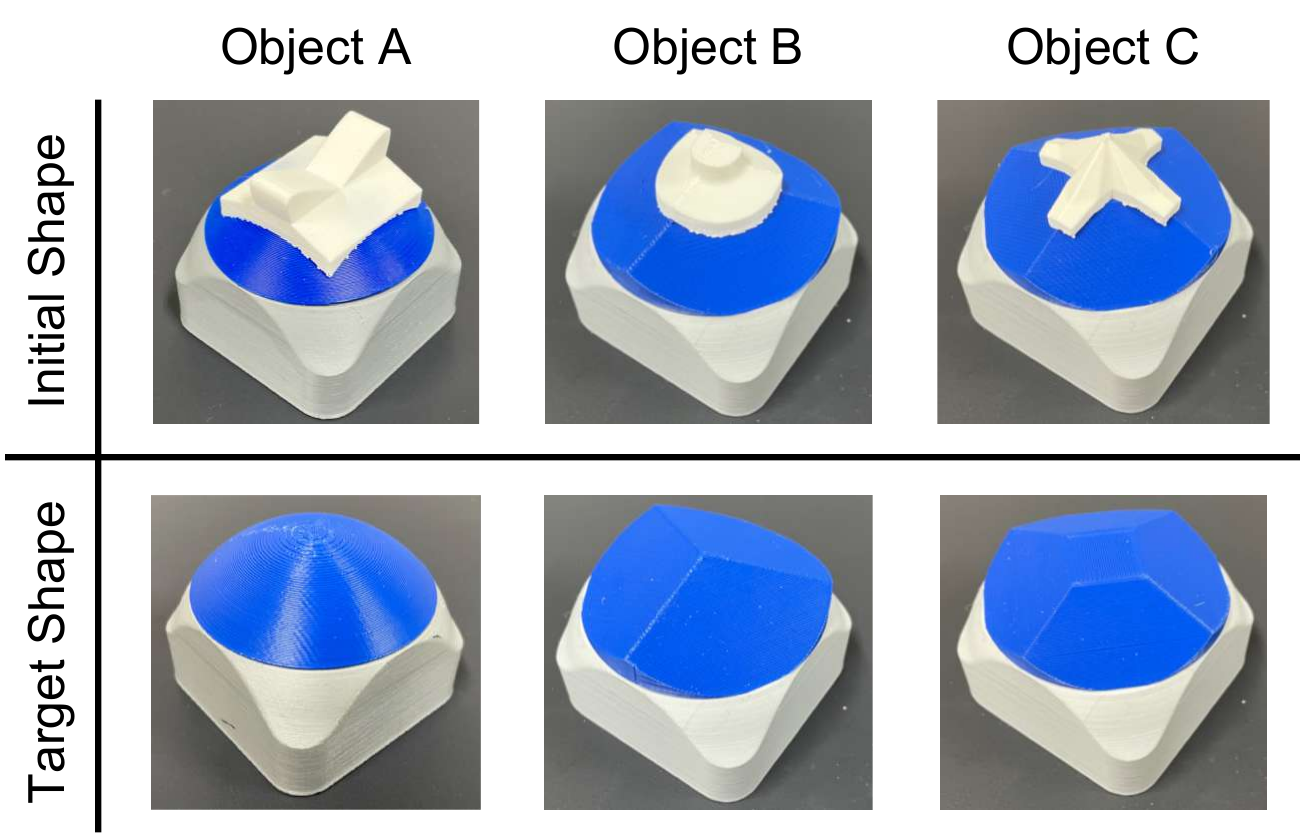}
    \caption{Initial and target shapes for experiments}
    \vspace{-3.0truemm}
    \label{fig:test_obj_mesh}
\end{figure}

\subsection{Iterative Optimization of CSDM and Cutting-Surface Sequence Planning by MBRL}\label{subsection:cutting_surface_aware_MBRL}
Finally, we formulate the iterative optimization of CSDM and MPC action planning (and execution) as an MBRL framework for high data efficiency as shown in 
\figurename\ref{fig:propose_method_large:c} and \algorithmcfname.\ref{algo:mbrl_alg}.
A total of \(T\) steps are defined as one learning episode of the RL process, and the total number of learning times is denoted as \(N_{episode}\).
In advance, we collected initial data using only GCM with \(N_{init}\) random actions and trained CSDM (L5-11).
The proposed framework follows an iterative procedure:
1) Execute MPC using CSAM. MPC minimizes the cost function and selects the next action (cutting surface) based on the observed shape in the task horizon \(t\). At each action step, we collect data to update CSDM (L16-22). 2) Train and update CSDM at the end of the episode (L23). By repeating this process \(N_{episode}\) times, the proposed framework learns CSDM and obtains the optimal cutting-surface sequence.

\section{Robotic Rough Grinding Environments}
This section describes our robotic rough grinding environment as shown in \figurename\ref{fig:eye_cath}. The grinding belt is fixed in the environment, and the object is mounted on the 6-DoF robot's end-effector. 
\figurename\ref{fig:test_obj_mesh} shows the three different initial and target shapes of the objects used for the experiment. These were all fabricated by a 3D printer. The gray part (bottom) is the base, the blue part (middle) is the target shape, and the white part (top) is the material to be removed by grinding. 

The robot action \({\mathbf{a}_t}\) manipulates the coordinates of the object around the work origin, which is an arbitrary distance away from the grinding belt, as equivalent to specifying the cutting surface of the object. 
In the grinding process using the grinding belt, we assume that the belt surface is sufficiently wide.
The 2-DoF of parallel translation and the 1-DoF rotation around a perpendicular axis can be ignored.
As a result, robot action becomes \(\mathbf{a}_t:=[\rm{roll}_t,\rm{pitch}_t,\rm{z}_t]\in\mathbb{R}^3\).
Since our MPC plans cutting surface at each time step, the planned cutting surface is executed by the following controllers: 1) Transition from the work origin to the target frame, 2) Holding the target frame for an arbitrary time, and 3) Returning to the work origin from the target frame.

\section{SIMULATION EXPERIMENT}
We evaluated the proposed method with a simulator having grinding dynamics. 
This experiment has three objectives:  
The first is to confirm the learning performance of our MBRL for grinding. 
The second is to evaluate the performance of optimized action sequences for grinding with the learned CSDM. 
We compared five comparison methods; details are described in Section \ref{subsub:comparison_method}. 
The third objective is to verify the applicability of the learned CSDM to different initial and target shapes.
For this purpose, the CSDM learned by Object A was applied to process Object B and to Object C and then evaluated.
The details of MBRL and MPC parameters used in experiments are on our project page\footnote{\url{https://t-hachimine.github.io/csambrl/}\label{fot:webpage}}.

\subsection{Simulation Environments} 
The initial and target shapes, modeling the same ones in \figurename\ref{fig:test_obj_mesh}, are created by CAD and converted to the point cloud.
Open3D \cite{Zhou2018} and Pyvista \cite{sullivan2019pyvista} are used for rendering.

To derive the deviation of the cutting surface, we implemented a virtual grinding resistance model, which depends on the robot action and removal volume at each time step.
In the actual grinding process, if the removal volume exceeds the physical limit, the grinding resistance is massively increased.
These conditions may lead to the destruction of the robot and the object.
Consequently, the robot interrupts the action when grinding resistance is applied above a certain threshold to the end-effector. 
To replicate this mechanism, the simulator sets a massive deviation of the cutting surface when the removal volume at one time step exceeds a threshold.

\subsection{Comparison Methods}\label{subsub:comparison_method}
To confirm the effectiveness of the proposed method, we prepared comparison methods as follows.
\textbf{Random}: selects actions randomly at each time step from a uniform distribution.
If the selected action exceeds the target shape, it is chosen again.
\textbf{Geometric}: performs MPC using only the GCM.
\textbf{Direct Shape Deformation Regression (DirectReg)}: performs MPC using a transition model that directly predicts the shape at the next step based on the current shape and robot action. The transition model is constructed with Point Cloud Auto Encoder (PCAE) using PointNet \cite{qi2017pointnet}.
{\textbf{MFRL-SAC}}: Performs Model-free RL by Soft Actor-Critic \cite{haarnoja2018soft}.
{\textbf{Proposed-GT}: MPC performs with ground truth cutting-surface deviation to confirm the upper bound of task performance in the proposed method.}

\subsection{MBRL and MPC Settings}
CSDM learns 1-axis deviation {\({y}_i=[\delta a_t]\)} of the cutting surface from input {\(\mathbf{x}_i := [\mathbf{e}_t,\mathbf{a}_t] = [V^{\rm geom}_t, {\rm{roll}}_t]\in \mathbb{R}^2\)}, where {\(V^{\rm geom}_t\)} is the removal volume, which is the sum of the Euclidean distance between the removal shape \(\mathbf{c}_{t+1}^{\rm geom}\) and the cutting surface divided by the number of removal shape point clouds.
We randomly collected 15 cutting surfaces with a removal volume of \([0.0\sim4.0]\) as initial data to train CSDM.
In training CSDM, we excluded from the training data that data obtained when the robot action was interrupted due to the removal volume exceeding a threshold value.
The ARD squared exponential kernel \cite{3569} is used as the GP kernel for CSDM.
MPC uses the Random Shooting method \cite{nagabandi2020deep,nagabandi2018neural} to minimize the cost function.
The over-cutting cost of \equationname\ref{eq:mpc_cost_func} returns 1000 if the predicted shape exceeds the target shape, and 0 otherwise.
The coefficient of variance cost \(\eta\) is set by the following equation:
\begin{equation}
\eta = 
\alpha \left[\frac{\footnotesize {d_{\rm CD}({\mathbf{s}}_{t},\mathbf{s}_{\rm{target}})}}
     {\footnotesize {d_{\rm CD}({\mathbf{s}}_{L,\rm{init}},\mathbf{s}_{L,\rm{target}})}}\right]^{\beta},
\label{eq:mpc_variance_coeff}
\end{equation}
where \(\alpha\) is the scale parameter of variance cost and \(\beta\) is the reduction factor based on the ratio of the initial shape error to the current shape error of the shape \(\mathbf{s}_{L}\) used in training.
The variance cost is affected by measurement noise in the real world, and thus it is decreased according to the ratio of the current shape error.

\begin{figure}[t!]
\begin{minipage}[t]{\columnwidth}
    \centering
    \includegraphics[clip,width=0.95\columnwidth]{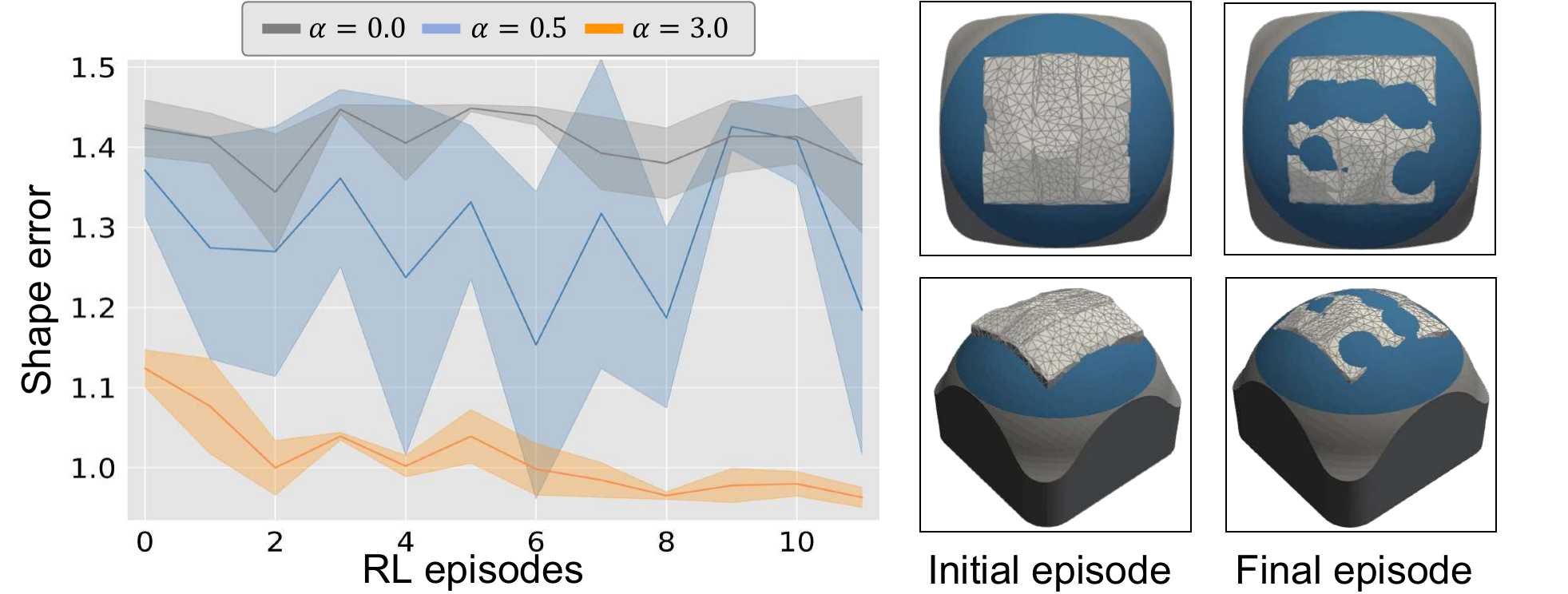}
    \caption{
    MBRL results for Object A in simulation. 
    \textbf{Left}: Learning curves, with mean value and standard deviation of three different initial data. 
    \textbf{Right}: Rendered shape before and after learning (\(\alpha=3.0\)).}
    \label{fig:sim_result:model_a_sim_rl_learn_curve}
\end{minipage}
\begin{minipage}[t]{\columnwidth}
    \centering
    \vspace{2.0truemm}
        \includegraphics[clip,width=\columnwidth]{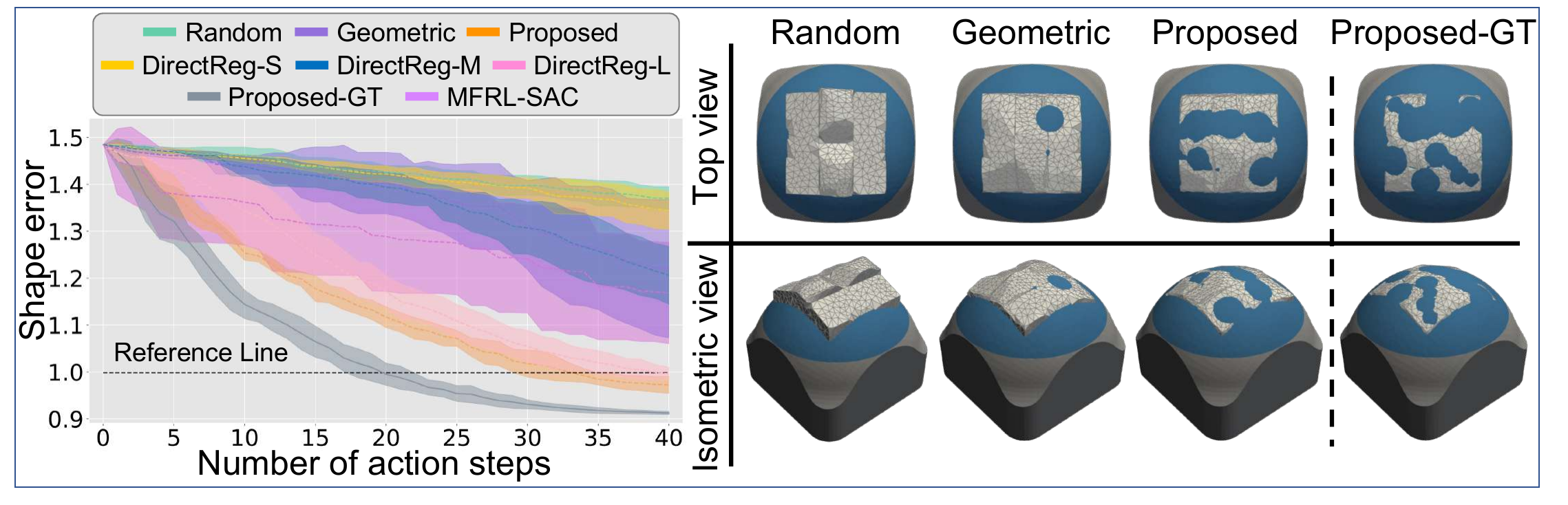}
    \begin{minipage}[b]{0.45\columnwidth}
        \centering
        \subcaption{Task performance}
    \label{fig:sim_result:model_a_task_performance}
    \end{minipage}
    \begin{minipage}[b]{0.53\columnwidth}
        \centering
        \subcaption{Shape examples of each method}
        \label{fig:sim_result:model_a_image_render}
    \end{minipage}
    \caption{
    Results of simulation experiments for Object A.
    (a) Task performance comparison for each method.
    (b) Rendered image of the shape at \(t=40\).}
    \label{fig:sim_result:model_a}
\end{minipage}
\begin{minipage}[t]{\columnwidth}
    \centering
    \vspace{2.0truemm}
        \includegraphics[clip,width=\columnwidth]{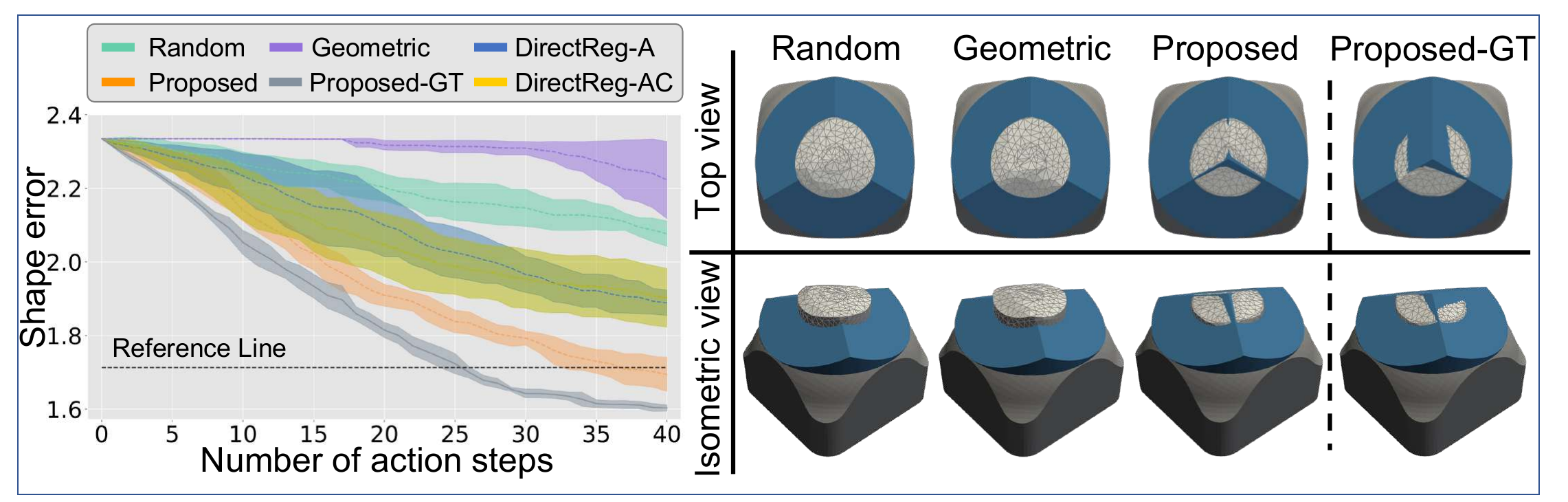}
    \begin{minipage}[b]{0.45\columnwidth}
        \centering
        \subcaption{Task performance}
    \end{minipage}
    \begin{minipage}[b]{0.53\columnwidth}
        \centering
        \subcaption{Shape examples of each method}
    \end{minipage}
    \caption{
    Results of simulation experiments for Object B. 
    (a) Task performance comparison for each method.
    (b) Rendered image of the shape at \(t=40\).}
    \label{fig:sim_result:model_b}
\end{minipage}
\begin{minipage}[t]{\columnwidth}
    \centering
    \vspace{2.0truemm}
        \includegraphics[clip,width=\columnwidth]{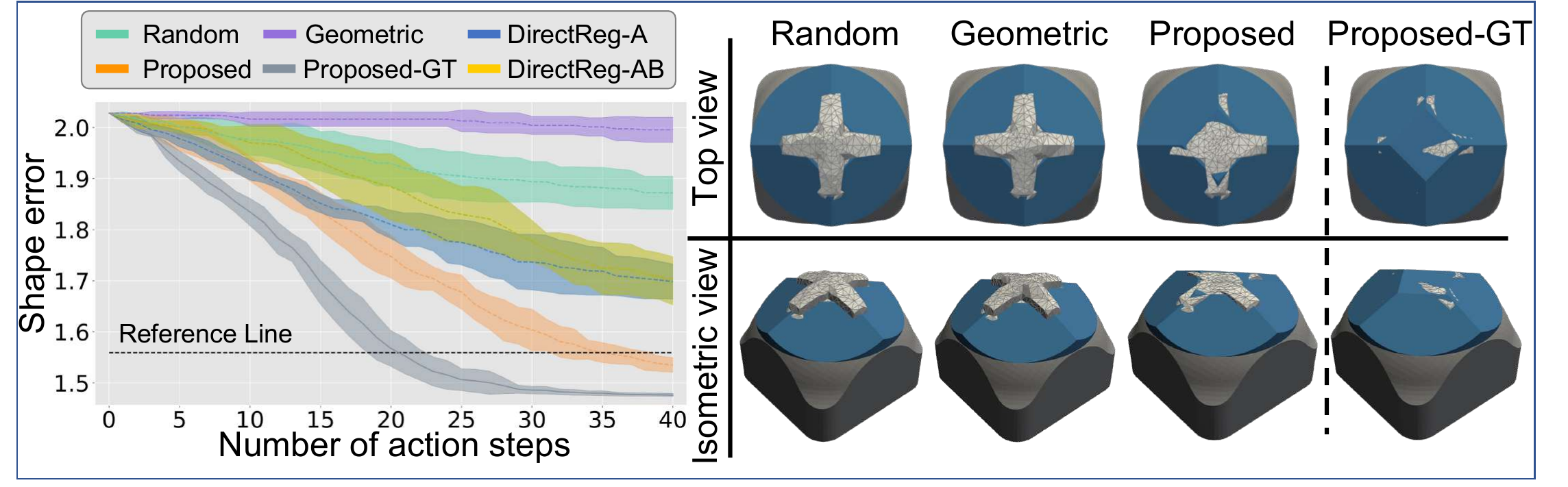}
    \begin{minipage}[b]{0.45\columnwidth}
        \centering
        \subcaption{Task performance}
    \end{minipage}
    \begin{minipage}[b]{0.53\columnwidth}
        \centering
        \subcaption{Shape examples of each method}
    \end{minipage}
    \caption{
    Results of simulation experiments for Object C. 
    (a) Task performance comparison for each method.
    (b) Rendered image of the shape at \(t=40\).}
    \label{fig:sim_result:model_c}
    \vspace{-3.0truemm}
\end{minipage}
\end{figure}

\subsection{Simulation Results}
\subsubsection{Learning Performance of Proposed MBRL}
\figurename\ref{fig:sim_result:model_a_sim_rl_learn_curve} shows the learning curves of MBRL for object A under different weights \(\alpha\) of the variance cost.
In the following experiment, the shape error is calculated by \equationname\ref{eq:champer_distaces}.
Note that \equationname\ref{eq:champer_distaces} is calculated with the Euclidean distance between the nearest neighbor points of two-point clouds, hence it is rare for the shape error to be zero unless two-point clouds are perfectly matched. These results confirm the ability of the proposed MBRL to optimize cutting-surface sequences with reduced shape error by setting the variance cost appropriately.

\subsubsection{Evaluation with Comparison Methods}
\figurename\ref{fig:sim_result:model_a} (a)
shows the task performance of each method.
\textbf{Proposed} uses each of the last two episodes of CSDM trained with three different initial data.
\textbf{DirectReg-S}, \textbf{DirectReg-M}, and \textbf{DirectReg-L} were trained with 4.5K, 91K, and 910K step data, respectively.
The Reference Line is defined at 15\(\, [\mathbin{\%}]\) above the average final shape error processed by \textbf{Proposed-GT}, for the indicator of task progress.
The graph shows that \textbf{Proposed} achieves the closest shape error to {\textbf{Proposed-GT}}.
\textbf{Geometric} does not incorporate the deviation of the cutting surface by grinding resistance, and thus the task performance is low.
In the case of \textbf{DeltaReg}, performance improves with increasing training data, but we need to collect more than 910K step data. Task performance of \textbf{MFRL-SAC} trained with 91K step data is comparable to \textbf{DirectReg-M}.
On the other hand, \textbf{Proposed} achieves the highest performance with only 0.5K step data.
These results confirm that the proposed method is more sample-efficient than model-free RL and direct shape prediction methods.
\figurename\ref{fig:sim_result:model_a} (b) shows an example of the shape after processing.

\tablename\ref{tab:sim_csam_error} shows the comparison of shape prediction error rate {\small\(\lvert d_{CD}\left({\hat{\mathbf{s}}_{t},{\mathbf{s}}_{target}}\right)-
d_{CD}\left({\mathbf{s}_{t},\mathbf{s}_{target}}\right)\rvert/
d_{CD}\left({\mathbf{s}_{t},\mathbf{s}_{target}}\right)\times100\)} and deviation of the cutting surface at the first step \(t=1\) for six times of each method, to evaluate it from the same shape state.
\textbf{Proposed} has a lower shape prediction error rate and cutting-surface deviation than \textbf{Geometric}.
The results suggest that learning CSDM improves the shape prediction accuracy of CSAM and allows the selection of actions with small cutting-surface deviation.

\subsubsection{Shape Generalization Performance}
We conducted six experiments on Objects B and C using CSDM learned for Object A. 
\textbf{Proposed} uses a final-episode
CSDM model that was trained with three different initial data
for Object A, and we conducted experiments on each twice.
\figurename\ref{fig:sim_result:model_b} and \figurename\ref{fig:sim_result:model_c} show comparisons of task performance for Object B and Object C, respectively.
\textbf{DirectReg-A} was trained with 91K step data using Object A, and \textbf{DirectReg-AB} and \textbf{AC} were trained with a total of 182K step data using Objects A and B, and Objects A and C, respectively. 
From \figurename\ref{fig:sim_result:model_b} and \figurename\ref{fig:sim_result:model_c}, it can be seen that \textbf{DeltaReg} task performance is lower than \textbf{Proposed}.
Therefore, a transition model with direct shape input and output would not provide sufficient shape generalization performance even if trained with multiple objects.

\begin{table}[!t]
\caption{
Comparison of shape prediction and cutting surface deviation for Object A at \(t=1\) in simulation experiments.
Here, \(*\), \(\star\), \(\dagger\) and \(\ddag\) denotes \(p< 0.05\) on paired t-test.}\label{tab:sim_csam_error}
\centering
\setlength{\tabcolsep}{5.5pt}
\scalebox{.86}{
\begin{tabular}{ccccc}
    \toprule
    {\textbf{Geometric}} &{\textbf{DirectReg-S}} & {\textbf{DirectReg-M}} &{\textbf{DirectReg-L}} & {\textbf{Proposed}} \\
    \midrule
    \multicolumn{5}{c}{Shape prediction error rate \(\, [\mathbin{\%}]\)} \\ 
    \midrule \midrule
        \(16.6\pm 2.7^{*}\) &\({25.1}\pm2.5^{\star}\) & \(20.0\pm 2.3^{\dagger}\)&\({13.1}\pm2.2^{\ddag}\)&\(\mathbf{0.16}\pm0.3^{*,\star,\dagger,\ddag}\)  \\
    \midrule
    \multicolumn{5}{c}{Cutting surface deviation \(\,\mathrm{[mm]}\)} \\ 
    \midrule \midrule
        \(7.0\pm1.8^{*}\)   &\({5.2}\pm2.2^{\star}\)& \(5.2\pm 2.7^{\dagger}\)&\({3.5}\pm1.8^{}\)& \(\mathbf{2.1}\pm0.76^{*,\star,\dagger}\) \\
    \bottomrule 
\end{tabular}}
\vspace{-3.0mm}
\end{table}

\section{REAL ROBOT EXPERIMENT}
\subsection{Construction of Robotic Rough Grinding System}
We use a 6-DoF robot (UR5e) equipped with a force and torque sensor.
In addition, a 3D vision sensor (YCAM3D-10M) is installed in the system to capture the object's shape. 
The robot moves in front of the 3D vision sensor and takes a point cloud.
The belt grinder (BDS-100) is equipped with a grinding belt having a grain size of WA\#80.
Details of our experimental system are on our project page\footref{fot:webpage}.

A hybrid position-force controller is implemented in the robot to prevent damage by grinding resistance.
Furthermore, the robot has a protection control that interrupts the action when a force or torque over the threshold value is applied to the robot's end-effector. This protection control is adopted to prevent damage to the robot when large grinding resistance is suddenly applied. 

\begin{figure}[t!]
\begin{minipage}[t]{\columnwidth}
    \centering 
        \includegraphics[width=0.95\columnwidth]{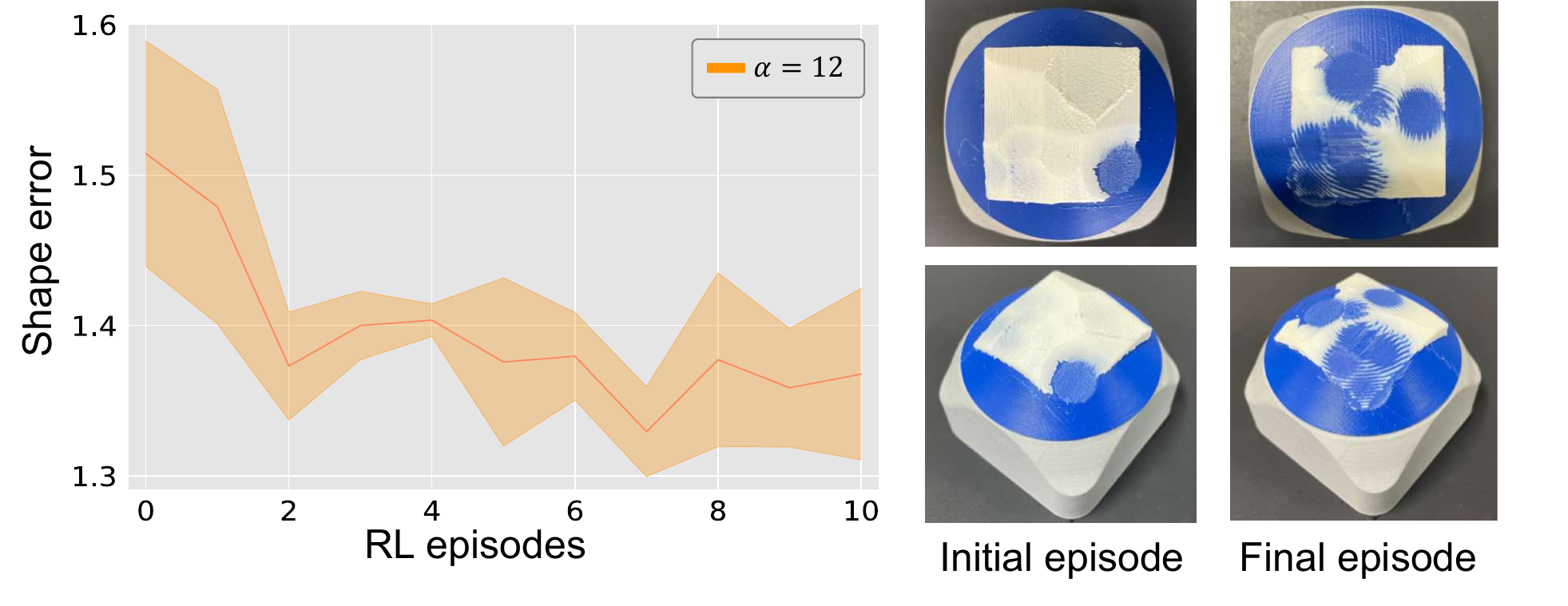}
        \caption{
        MBRL results for Object A in robot experiments. 
        \textbf{Left}: Learning curves, with mean value and standard deviation of three different initial data. 
        \textbf{Right}: Captured shape before and after learning (\(\alpha=12.0\)).}
        \label{fig:real_result:rl_leaning_curve}
\end{minipage}
\begin{minipage}[t]{\columnwidth}
    \centering
   \vspace{2.0truemm}
        \includegraphics[clip,width=\columnwidth]{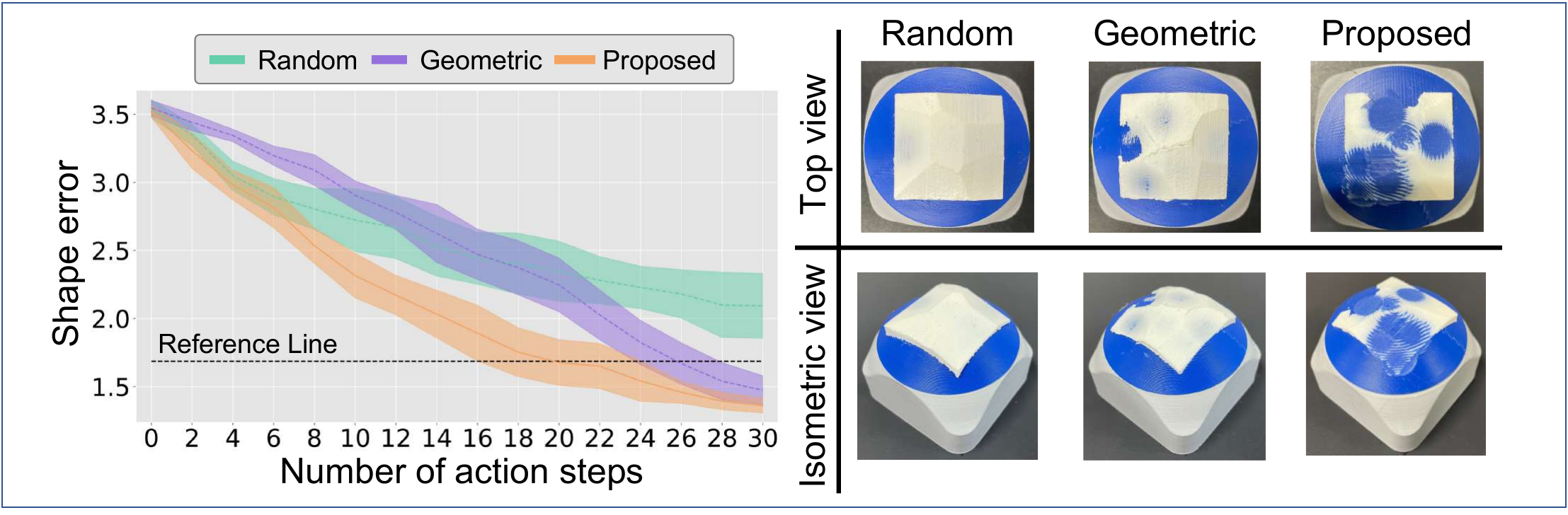}
    \begin{minipage}[b]{0.5\columnwidth}
        \centering
        \subcaption{Task performance }
        \label{fig:real_result:model_a_task_performance}
    \end{minipage}
    \begin{minipage}[b]{0.48\columnwidth}
        \centering
        \subcaption{Shape examples of each method}
        \label{fig:real_result:model_a_render_image}
    \end{minipage}
    \caption{Results of robot experiments for Object A. 
    (a) Task performance comparison for each method.
    (b) Captured image of the shape at \(t=30\).}
    \label{fig:real_result:model_a}
\end{minipage}
\begin{minipage}[t]{\columnwidth}
    \centering
   \vspace{2.0truemm}
    \includegraphics[clip,width=\columnwidth]{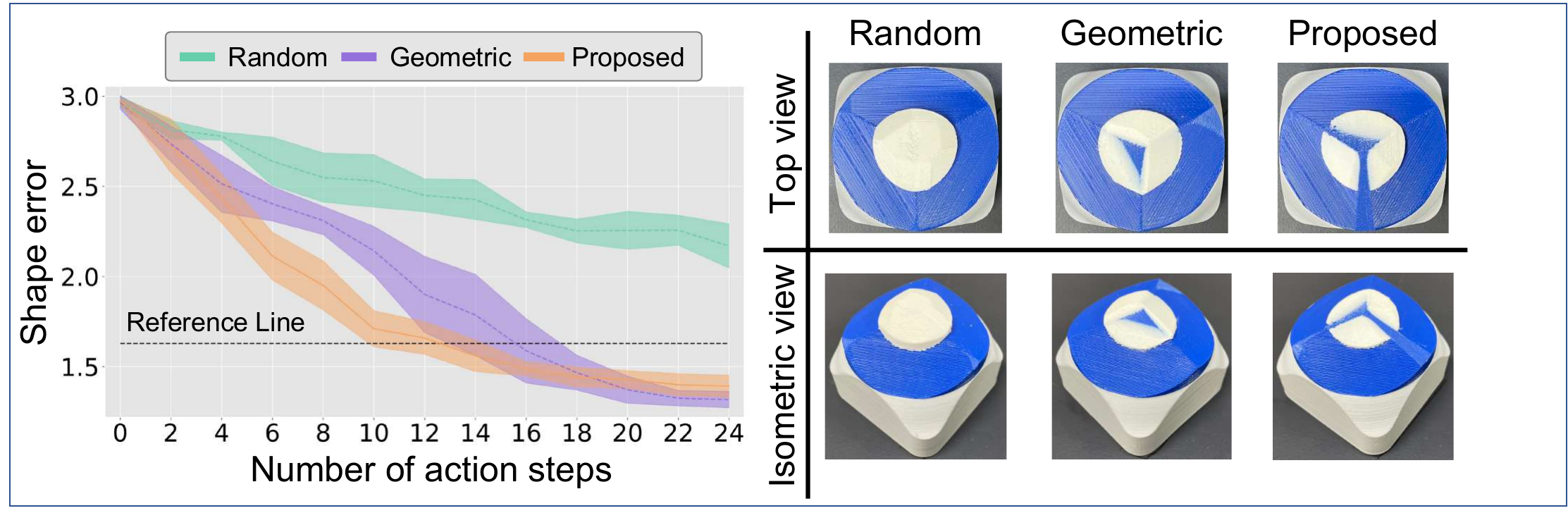}
    \begin{minipage}[b]{0.5\columnwidth}
        \centering
        \subcaption{Task performance}
        \label{fig:real_result:model_b_trend}
    \end{minipage}
    \begin{minipage}[b]{0.48\columnwidth}
        \centering
        \subcaption{Shape examples of each method}
        \label{fig:real_result:model_b_render_image}
    \end{minipage}
    \caption{Results of robot experiments for Object B.
    (a) Task performance comparison for each method.
    (b) Captured image of the shape at \(t=14\).}
    \label{fig:real_result:model_b_result}
\end{minipage}
\begin{minipage}[t]{\columnwidth}
    \centering
   \vspace{2.0truemm}
    \includegraphics[clip,width=\columnwidth]{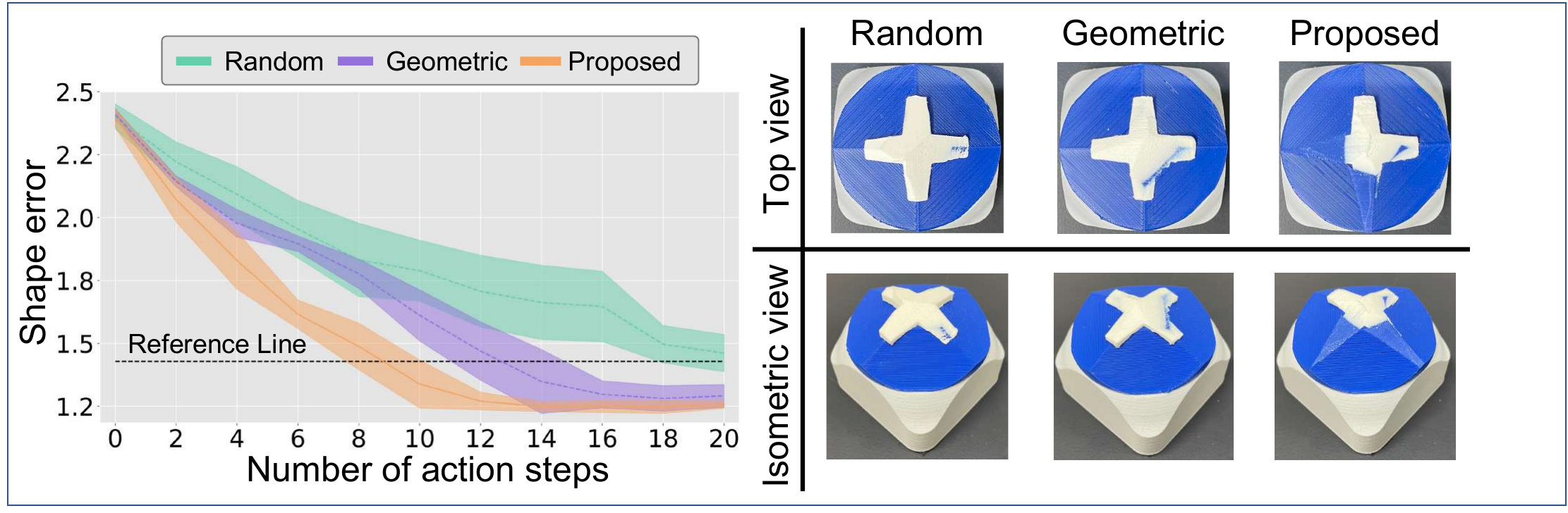}
    \begin{minipage}[b]{0.5\columnwidth}
        \centering
        \subcaption{Task performance}
        \label{fig:real_result:model_c_trend}
    \end{minipage}
    \begin{minipage}[b]{0.48\columnwidth}
        \centering
        \subcaption{Shape examples of each method}
        \label{fig:real_result:model_c_render_image}
    \end{minipage}
    \caption{Results of robot experiments for Object C. 
    (a) Task performance comparison for each method.
    (b) Captured image of the shape at \(t=10\).}
    \label{fig:real_result:model_c_result}
\end{minipage}
\end{figure}

\subsection{MBRL and MPC Settings}
CSDM learns 1-axis deviation {{\({y}_i=[\delta a_t]\)}} of the cutting surface from input {{\(\mathbf{x}_i := [\mathbf{e}_t,\mathbf{a}_t] = [{V}^{\rm geom}_t,{\rm w}_t,{\rm h}_t,{\rm d}_t,{\rm roll}_t,{\rm pitch}_t]\in \mathbb{R}^6\)}}, where \({\rm w}_t,{\rm h}_t,{\rm d}_t\) is the 3D-bounding box of the removal shape.
In the actual grinding process, the grinding resistance depends on the contact area to the grinding belt, even if the removal volume at the time step is the same.
Therefore, the input dimension of CSDM is extended.
We randomly collected 60 cutting surfaces
with a removal volume of \([0.1\sim2.0]\) as initial data to train CSDM.
In training CSDM, the data obtained when the robot action was interrupted by protective control were excluded. 
The MPC cost function follows \equationname\ref{eq:mpc_cost_func} and its minimization method is the same as the simulation.
\begin{table}[t!]
    \centering
    \caption{Comparison of the number of time steps to achieve Reference Line. Here, \(*\) denotes \(p< 0.05\) on paired t-test.}\label{tab:task_success_compare}
    \begin{tabular}{l|c|c}
    \toprule
     & \textbf{Geometric} & \textbf{Proposed} \\ 
    \midrule
    {Object A}&\(27.3\pm1.9 ^{*}\)&\(\mathbf{20.0}\pm{{3.1}}^{*}\) \\
    {Object B}&\(16.7\pm2.5^{*}\)&\(\mathbf{13.0}\pm{{2.0}}^{*}\)  \\ 
    {Object C}&\(14.0\pm2.3^{*}\)&\(\mathbf{9.7}\pm{{1.4}}^{*}\) \\ 
    \bottomrule
    \end{tabular}
    \vspace{-3.8truemm}
\end{table}

\begin{table}[t]
\caption{Comparison of shape prediction and cutting surface deviation for Object A at \(t=1\) in robot experiments. 
Here, * denotes \(p< 0.05\) on paired t-test.}\label{tab:comp_action_proper}
\centering
\setlength{\tabcolsep}{5pt}
\scalebox{1.}{
\begin{tabular}{l|c|c}
    \toprule
        & {\textbf{Geometric}} & {\textbf{Proposed}} \\ 
     \midrule 
        Shape prediction error rate\(\, [\mathbin{\%}]\)& \(27.0\pm 5.37^{*}\)&\(\mathbf{5.10}\pm2.32^{*}\) \\
        Cutting surface deviation\(\,\mathrm{[mm]}\)& \(6.85\pm1.30^{*}\)&\(\mathbf{1.37}\pm1.86^{*}\)\\
    \bottomrule
\end{tabular}
}
\end{table}

\begin{figure}[!t]
    \centering
    \includegraphics[clip,width=1.0\columnwidth]{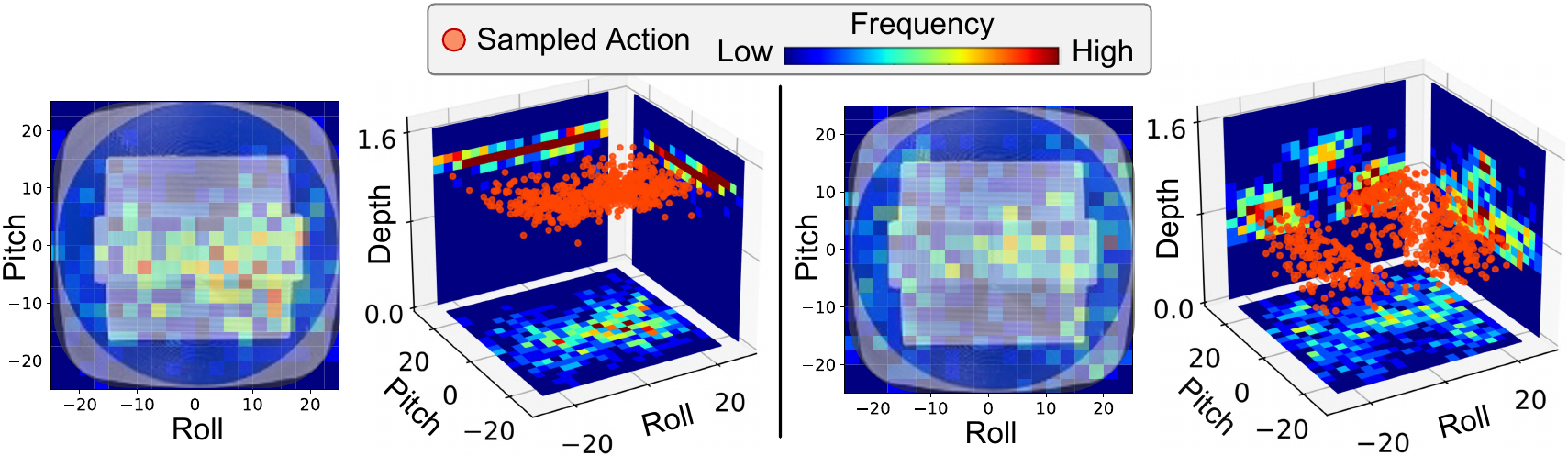}
        \begin{minipage}[b]{0.5\columnwidth}
        \centering
        \subcaption{Object A with \textbf{Geometric}}
    \end{minipage}
    \begin{minipage}[b]{0.48\columnwidth}
    \centering
        \subcaption{Object A with \textbf{Proposed}}
    \end{minipage}
    \caption{Distribution of cutting surfaces sampled 600 times for the first step of Object A}
    \label{fig:real_result:compare_action}
    \vspace{-3.5truemm}
\end{figure}

\subsection{Results}
\subsubsection{Learning Performance of Proposed MBRL}
\figurename\ref{fig:real_result:rl_leaning_curve} shows the learning curve of our MBRL for Object A.
These results confirm that, similar to the simulation results, the proposed MBRL can optimize the cutting-surface sequence with reduced shape error in a real environment.
\subsubsection{Evaluation with Comparison Methods}
\figurename\ref{fig:real_result:model_a} (a) shows the task performance of each method.
\textbf{Proposed} uses each of the last two episodes of CSDM trained with three different initial data.
The settings of \textbf{Random} and \textbf{Geometric} are the same as in the Simulation.
Reference Line is defined at 15\(\, [\mathbin{\%}]\) above the average final shape error processed by \textbf{Proposed} because the ground truth cutting-surface deviation is unknown.
In real robot experiments, the action space is limited compared to the simulation because the action is sometimes constrained based on the force sensor for safety reasons.
\figurename\ref{fig:real_result:model_a} (a) shows that \textbf{Proposed} achieves Reference Line more quickly than \textbf{Geometric}. 
\tablename\ref{tab:task_success_compare} compares the number of time steps to achieve the Reference Line between \textbf{Proposed} and \textbf{Geometric}.
The paired t-test results show that the task performance of \textbf{Proposed} is significantly different from that of \textbf{Geometric}.
\figurename\ref{fig:real_result:model_a} (b) shows an example of the shape after processing.

\tablename\ref{tab:comp_action_proper} shows the comparison of shape prediction error rate and deviation of the cutting surface at the first step \(t=1\) for six times of each method, to evaluate it from the same shape state.  
\textbf{Proposed} has a lower prediction shape error rate and cutting-surface deviation than \textbf{Geometric}.
In addition, \figurename\ref{fig:real_result:compare_action} shows the distribution of the cutting surface at \(t=1\).
This distribution shows that \textbf{Proposed} selects the cutting surfaces with a small \(\rm z_{t}\) (Depth) and large \(\rm roll_{t}\) (Roll) and \(\rm pitch_{t}\) (Pitch) compared to \textbf{Geometric}. 
These results suggest that learning CSDM improves the shape prediction accuracy of CSAM and allows the selection of actions with small cutting-surface deviation by optimizing the depth and angle of contact with the grinding belt.

\subsubsection{Shape Generalization Performance}
We conducted six experiments on Objects B and C using CSDM learned
for Object A.
\textbf{Proposed} uses a final-episode CSDM model trained with three different initial data for Object A, and we conduct experiments on each twice.
\tablename\ref{tab:task_success_compare} compares the number of time steps to achieve Reference Line.
The paired t-test results show that the task performance of \textbf{Proposed} significantly differs from that of \textbf{Geometric}.
\figurename\ref{fig:real_result:model_b_result} and \figurename\ref{fig:real_result:model_c_result} show the task performance and the actual shape after processing by each method.

\section{DISCUSSION}
This study focused on adaptation to the action-dependent process conditions in robotic grinding, assuming that environment-dependent process conditions are fixed. Therefore, we must retrain CSDM if the material hardness or grinding belt is changed. To achieve this, we will embed environment-dependent process conditions as parameters in the CSDM dataset, which is expected to provide, in future work, applicability to a wider variety of materials and grinding belts.

A limitation of the proposed method is that the robot motion is specified as a cutting surface, which may lead to inefficient processing for smooth shapes such as round surfaces. A possible solution to this problem is to extend the cost function to account for the continuity of the trajectory, but a multi-objective cost function may not be desirable from an optimization perspective. Therefore, learning a cost function that includes trajectory continuity based on human demonstrations \cite{admm} is an interesting future challenge.
In addition, the trade-off between the computational cost of GCM and the quality of the shape representation could be a bottleneck for real-time action planning with MPC. We believe computational efficiency can be improved by implementing GCM with parallel computing.

\section{CONCLUSION}
We proposed the MBRL method for object shaping through removal processing by grinding. 
The key idea of the proposed method is to represent the shape-transition model by GCM, which is independent of process conditions, and CSDM, which only depends on process conditions, by introducing the {cutting surface}.
Experiments were conducted on simulators and in the real world to evaluate the effectiveness of the proposed method.
The experimental results confirmed that the proposed method achieves efficient learning with a reasonable amount of data and provides generalization capability for initial and target shapes that differ from the training data.

\bibliographystyle{ieeetr}
\bibliography{reference}

\begin{thebibliography}{10}

\bibitem{sanchez2018robotic}
J.~Sanchez, J.~Corrales, B.~Bouzgarrou, and Y.~Mezouar, ``Robotic manipulation
  and sensing of deformable objects in domestic and industrial applications: a
  survey,'' {\em The International Journal of Robotics Research}, vol.~37,
  no.~7, pp.~688--716, 2018.

\bibitem{9341102}
H.~Yang, C.~Shih, Y.~Lo, and F.~Lian, ``Zero-tuning grinding process
  methodology of cyber-physical robot system,'' in {\em 2020 IEEE/RSJ
  International Conference on Intelligent Robots and Systems}, pp.~4270--4275,
  2020.

\bibitem{matl2021deformable}
C.~Matl and R.~Bajcsy, ``Deformable elasto-plastic object shaping using an
  elastic hand and model-based reinforcement learning,'' in {\em 2021 IEEE/RSJ
  International Conference on Intelligent Robots and Systems}, pp.~3955--3962,
  2021.

\bibitem{nagabandi2020deep}
A.~Nagabandi, K.~Konolige, S.~Levine, and V.~Kumar, ``Deep dynamics models for
  learning dexterous manipulation,'' in {\em Conference on Robot Learning},
  pp.~1101--1112, 2020.

\bibitem{deisenroth2011pilco}
M.~Deisenroth and C.~E. Rasmussen, ``Pilco: A model-based and data-efficient
  approach to policy search,'' in {\em Proceedings of the 28th International
  Conference on machine learning}, pp.~465--472, 2011.

\bibitem{TANG20092847}
J.~Tang, J.~Du, and Y.~Chen, ``Modeling and experimental study of grinding
  forces in surface grinding,'' {\em Journal of Materials Processing
  Technology}, vol.~209, no.~6, pp.~2847--2854, 2009.

\bibitem{8403315}
T.~Tang, C.~Wang, and M.~Tomizuka, ``A framework for manipulating deformable
  linear objects by coherent point drift,'' {\em IEEE Robotics and Automation
  Letters}, vol.~3, no.~4, pp.~3426--3433, 2018.

\bibitem{9197121}
P.~Sundaresan, J.~Grannen, B.~Thananjeyan, A.~Balakrishna, M.~Laskey, K.~Stone,
  J.~E. Gonzalez, and K.~Goldberg, ``Learning rope manipulation policies using
  dense object descriptors trained on synthetic depth data,'' in {\em 2020 IEEE
  International Conference on Robotics and Automation}, pp.~9411--9418, 2020.

\bibitem{9001169}
D.~S^^c3^^a1nchez, W.~Wan, and K.~Harada, ``Tethered tool manipulation planning
  with cable maneuvering,'' {\em IEEE Robotics and Automation Letters}, vol.~5,
  no.~2, pp.~2777--2784, 2020.

\bibitem{9561391}
D.~Seita, P.~Florence, J.~Tompson, E.~Coumans, V.~Sindhwani, K.~Goldberg, and
  A.~Zeng, ``Learning to rearrange deformable cables, fabrics, and bags with
  goal-conditioned transporter networks,'' in {\em IEEE International
  Conference on Robotics and Automation}, pp.~4568--4575, 2021.

\bibitem{corl2020softgym}
X.~Lin, Y.~Wang, J.~Olkin, and D.~Held, ``Softgym: Benchmarking deep
  reinforcement learning for deformable object manipulation,'' in {\em
  Conference on Robot Learning}, pp.~432--448, 2021.

\bibitem{TSURUMINE201972}
Y.~Tsurumine, Y.~Cui, E.~Uchibe, and T.~Matsubara, ``Deep reinforcement
  learning with smooth policy update: Application to robotic cloth
  manipulation,'' {\em Robotics and Autonomous Systems}, vol.~112, pp.~72--83,
  2019.

\bibitem{inproceedings_meshcut}
E.~Sifakis, K.~Der, and R.~Fedkiw, ``Arbitrary cutting of deformable
  tetrahedralized objects,'' in {\em In Proceedings of the 2007 ACM
  SIGGRAPH/Eurographics symposium on Computer animation}, pp.~73--80, 2007.

\bibitem{heiden2021disect}
E.~Heiden, M.~Macklin, Y.~S. Narang, D.~Fox, A.~Garg, and F.~Ramos, ``{DiSECt:
  A Differentiable Simulation Engine for Autonomous Robotic Cutting},'' in {\em
  Proceedings of Robotics: Science and Systems}, 2021.

\bibitem{pomerleau1991efficient}
D.~A. Pomerleau, ``{Efficient Training of Artificial Neural Networks for
  Autonomous Navigation},'' {\em Neural computation}, vol.~3, no.~1,
  pp.~88--97, 1991.

\bibitem{hu2019difftaichi}
Y.~Hu, L.~Anderson, T.~Li, Q.~Sun, N.~Carr, J.~{Ragan-Kelley}, and F.~Durand,
  ``Difftaichi: Differentiable programming for physical simulation,'' {\em
  arXiv:1910.00935}, 2020.

\bibitem{kumaresan2000linear}
S.~Kumaresan, {\em Linear algebra: A geometric approach}.
\newblock PHI Learning Pvt. Ltd., 2000.

\bibitem{dunn20113d}
F.~Dunn and I.~Parberry, {\em 3D math primer for graphics and game
  development}.
\newblock CRC Press, 2011.

\bibitem{rowe2013principles}
W.~B. Rowe, {\em Principles of modern grinding technology}.
\newblock William Andrew, 2013.

\bibitem{3569}
C.~Rasmussen and C.~Williams, {\em Gaussian Processes for Machine Learning}.
\newblock Adaptive Computation and Machine Learning, Cambridge, MA, USA: MIT
  Press, 2006.

\bibitem{Nguyen2021PointsetDF}
T.~Nguyen, Q.~Pham, T.~Le, T.~Pham, N.~Ho, and B.~Hua, ``Point-set distances
  for learning representations of 3d point clouds,'' {\em 2021 IEEE/CVF
  International Conference on Computer Vision}, pp.~10458--10467, 2021.

\bibitem{Zhou2018}
Q.~Zhou, J.~Park, and V.~Koltun, ``{Open3D}: {A} modern library for {3D} data
  processing,'' {\em arXiv:1801.09847}, 2018.

\bibitem{sullivan2019pyvista}
C.~B. Sullivan and A.~Kaszynski, ``{PyVista}: 3d plotting and mesh analysis
  through a streamlined interface for the visualization toolkit ({VTK}),'' {\em
  Journal of Open Source Software}, vol.~4, no.~37, p.~1450, 2019.

\bibitem{qi2017pointnet}
C.~R. Qi, H.~Su, K.~Mo, and L.~J. Guibas, ``Pointnet: Deep learning on point
  sets for 3d classification and segmentation,'' in {\em Proceedings of the
  IEEE conference on computer vision and pattern recognition}, pp.~652--660,
  2017.

\bibitem{haarnoja2018soft}
T.~Haarnoja, A.~Zhou, P.~Abbeel, and S.~Levine, ``Soft actor-critic: Off-policy
  maximum entropy deep reinforcement learning with a stochastic actor,'' in
  {\em International conference on machine learning}, pp.~1861--1870, 2018.

\bibitem{nagabandi2018neural}
A.~Nagabandi, G.~Kahn, R.~S. Fearing, and S.~Levine, ``Neural network dynamics
  for model-based deep reinforcement learning with model-free fine-tuning,'' in
  {\em 2018 IEEE international conference on robotics and automation},
  pp.~7559--7566, 2018.

\bibitem{admm}
A.~Escontrela, X.~B. Peng, W.~Yu, T.~Zhang, A.~Iscen, K.~Goldberg, and
  P.~Abbeel, ``Adversarial motion priors make good substitutes for complex
  reward functions,'' in {\em 2022 IEEE/RSJ International Conference on
  Intelligent Robots and Systems}, pp.~25--32, 2022.

\end{thebibliography}

\end{document}